\documentclass{tlp}
\usepackage{aopmath}
\usepackage{graphicx} 

\submitted{2012.6.10}
\revised{2012.12.11}
\accepted{2013.3.8}

\newcommand{\expl}{\mbox{\rm{expl}}}

\newcommand{\ddefined}{\stackrel{{\rm def}}{=}}

\newcommand{\mswz}{\mbox{\tt msw}}                  
\newcommand{\db}{D\!B}
\newcommand{\pdb}{P_{D\!B}}
\newcommand{\pmsw}{P_{\tt msw}}

\newcommand{\btheta}{\mbox{\boldmath $\theta$}}

\newcommand{\bG}{\mbox{\bf{\em G}}}

\begin{document}

\long\def\comment#1{}

\title{  Viterbi training in PRISM  }

\author[T.Sato and K.Kubota]
{ Taisuke Sato \\
  Tokyo Institute of Technology \\
  E-mail: sato@mi.cs.titech.ac.jp
\and
  Keiichi Kubota\\
  Tokyo Institute of Technology \\
  E-mail: kubota@mi.cs.titech.ac.jp
}

\pagerange{\pageref{firstpage}--\pageref{lastpage}}
\volume{\textbf{10} (3):}
\jdate{March 2002}
\setcounter{page}{1}
\pubyear{2002}

\maketitle

\label{firstpage}

\begin{abstract} 
VT (Viterbi training),  or hard EM, is an efficient  way of parameter learning
for probabilistic models with hidden  variables.  Given an observation $y$, it
searches  for a  state  of hidden  variables  $x$ that  maximizes $p(x,y  \mid
\theta)$ by coordinate  ascent on parameters $\theta$ and  $x$.  In this paper
we  introduce VT  to PRISM,  a logic-based  probabilistic modeling  system for
generative  models.  VT  improves  PRISM in  three  ways.  First  VT in  PRISM
converges  faster than  EM in  PRISM due  to the  VT's  termination condition.
Second,  parameters  learned by  VT  often  show  good prediction  performance
compared to  those learned by EM.   We conducted two  parsing experiments with
probabilistic  grammars while learning  parameters by  a variety  of inference
methods, i.e.\ VT,  EM, MAP and VB.   The result is that VT  achieved the best
parsing accuracy among them in  both experiments.  Also we conducted a similar
experiment  for  classification  tasks  where  a  hidden  variable  is  not  a
prediction target unlike probabilistic grammars.  We found that in such a case
VT does  not necessarily  yield superior performance.   Third since  VT always
deals with a single probability  of a single explanation, Viterbi explanation,
the  exclusiveness condition  that is  imposed on  PRISM programs  is  no more
required if we learn parameters by VT.

Last but not least we can say that as VT in PRISM is general and applicable to
any  PRISM program, it  largely reduces  the need  for the  user to  develop a
specific VT algorithm for a specific model.  Furthermore since VT in PRISM can
be  used just  by  setting a  PRISM  flag appropriately,  it  makes VT  easily
accessible to (probabilistic) logic programmers.
To appear in Theory and Practice of Logic Programming (TPLP).
\end{abstract} 

\section{Introduction}
\label{intro}
VT (Viterbi  training) has been used  for long time as  an efficient parameter
learning  method  in  various  research  fields such  as  machine  translation
\cite{Brown93},  speech  recognition  \cite{Juang90,Strom99},  image  analysis
\cite{Joshi06},     parsing    \cite{Spitkovsky10}     and     gene    finding
\cite{Lomsadze05}.   Although  VT is  NP-hard  even  for PCFGs  (probabilistic
context free  grammars), which  is proved by  encoding the 3-SAT  problem into
PCFGs \cite{Cohen10},  and is  biased unlike MLE
(maximum likelihood estimation)\cite{Lember07},  it often outperforms and runs
faster than the conventional EM algorithm.

We introduce  this VT to  PRISM which is  a probabilistic extension  of Prolog
\cite{Sato01g,Sato08a}\footnote{
VT  is  available  in PRISM2.1.   PRISM2.1  is  the  latest version  of  PRISM
downloadable from {\tt http://sato-www.cs.titech.ac.jp/prism/}.
}.  There are already multiple  parameter learning methods available in PRISM.
One  is  the  EM algorithm,  or  more  generally  MAP (maximum  a  posteriori)
estimation \cite{Sato01g}.   Another is VB  (variational Bayes) \cite{Sato09a}
which  approximately  realizes Bayesian  inference  and  learns pseudo  counts
assuming Dirichlet  priors over parameters.   They are implemented  on PRISM's
data structure  called {\em explanation graph\/}s  representing AND/OR boolean
formulas made up of probabilistic ground atoms.  Probabilities used in EM, MAP
and VB  are all computed  by running the generalized  inside-outside algorithm
\cite{Sato01g} or its variant on explanation graphs.

VT in  PRISM runs on explanation  graphs just like  EM, MAP and VB  but always
deals with  a single probability of  a single explanation  called {\em Viterbi
  explanation\/} or  most probable explanation.   Compared to EM  that updates
only  parameters,   VT  alternately   updates  the  Viterbi   explanation  and
parameters, computing  one from  the other and  vice versa, until  the Viterbi
explanation stops changing.  Note that  this results in earlier termination of
the  algorithm than EM  because a  small perturbation  in parameters  does not
change the Viterbi explanation whereas it keeps EM running.  Actually we found
in  our experiments  in Section~\ref{sec:vt-prism}  that EM  required 8  to 15
times more cycles  to stop than VT.  Also since VT  updates parameters so that
they maximize the  probability of the Viterbi explanation,  it is possible and
probable that  the final  parameters by  VT give a  higher probability  to the
Viterbi explanation than  those learned by EM, which  intuitively explains why
VT  tends to  yield superior  performance to  EM in  prediction tasks  such as
parsing that computes the Viterbi explanation  as a predicted value, as we see
in Section~\ref{sec:grammar}.

In addition  VT brings about  a favorable side  effect on PRISM.  VT  does not
require {\em the exclusiveness condition\/} which is imposed on PRISM programs
to ensure  efficient sum-product probability computation.  This  is because VT
always deals with a single  probability of Viterbi explanation and hence there
is no  need for  summing up probabilities  of the  non-exclusive explanations.
Consequently PRISM can learn parameters by VT for programs that do not satisfy
the  exclusiveness condition.  We  will discuss  more about  the exclusiveness
condition in Section~\ref{sec:exclusivness}.

VT thus improves PRISM in the following points:
\begin{itemize}
\item Faster convergence due to a less number of iterations compared to EM
\item Ability to learn parameters good for prediction
\item The elimination of the exclusiveness condition imposed on programs
\end{itemize}

From  the viewpoint  of statistical  machine learning  and  PLP (probabilistic
logic programming), on the other hand, we can first say that PRISM generalizes
VT.  That  is, the  VT  algorithm implemented  in  PRISM  works for  arbitrary
probabilistic models described by PRISM, a Turing complete language, including
BNs  (Bayesian networks),  HMMs (hidden  Markov models)  and PCFGs,  and hence
eliminates  the need  for  the user  to  derive and  implement  a specific  VT
algorithm for a specific model that can be described as a PRISM program.  Also
it makes VT easily accessible  to probabilistic logic programmers because they
can  use  VT  just  by  setting  {\tt  learn\_mode},  one  of  PRISM's  flags,
appropriately.  As  a result, by  switching the {\tt learn\_mode}  flag he/she
can choose the  best parameter learning method for their  models from EM, MAP,
VB and  VT, all  available in PRISM2.1,  without rewriting and  adapting their
programs  to   each  parameter   learning  method.   Indeed,   the  exhaustive
comparisons among EM, MAP, VB and VT done in our experiments seem quite costly
in other environments.

In what  follows, we first review PRISM  in Section~\ref{sec:review-prism} and
then  explain  the  basic  idea  of   VT  and  reformulate  it  for  PRISM  in
Section~\ref{sec:vt-prism}.  We then apply VT to two probabilistic grammars in
Section~\ref{sec:grammar} using  the ATR corpus  where a hidden variable  in a
model  is a  prediction  target.   In Section~\ref{sec:nbh},  we  deal with  a
different situation  using an NBH (naive  Bayes with a  hidden variable) model
whose  hidden variable is  {\em not\/}  a prediction  target.  We  explain the
implication     of     VT     on     the    exclusiveness     condition     in
Section~\ref{sec:exclusivness}.      Section~\ref{sec:discussion}    discusses
related work.  Section~\ref{sec:conclusion} is the conclusion.

\section{Reviewing PRISM} 
\label{sec:review-prism}
For  the  self-containedness  we  review  PRISM focusing  on  its  computation
mechanism.  PRISM  is one of the  SRL (statistical relational  learning) / PLL
(probabilistic  logic learning) languages \cite{Getoor07,DeRaedt08b}  which aim
at using rich expressions such  as relations and first-order logic for complex
probabilistic modeling.   It is a  probabilistic extension of  Prolog enhanced
with  various built-in  predicates for  statistical machine  learning  such as
predicates  for parameter  learning,  Viterbi inference,  model scoring,  MCMC
sampling and so on in addition to standard predicates equipped with Prolog.

\subsection{Probability naively computed} 
Syntactically a PRISM  program $\db$ looks like a  usual Prolog program except
the use  of probabilistic  built-in predicate of  the form  {\tt msw($i$,$v$)}
called ``multi-valued random switch''(with  switch name $i$) that represents a
probabilistic  choice  using  simple  probabilistic events  such  as  dice
throwing; {\tt  msw($i$,$v$)} says  that throwing a  dice named $i$  yields an
outcome  $v$.  Let  $V_i =\{v_1,\ldots,v_{|V_i|}  \}$ be  the set  of possible
outcomes  for  $i$.   The  set  ${\tt  msw}(i,\cdot)  \ddefined  \{  \mbox{\tt
  msw($i$,$v$)}  \mid v  \in  V_i \}$  of {\tt  msw}  atoms is  given a  joint
distribution  such   that  one  of   the  ${\tt  msw}(i,\cdot)$'s,   say  {\tt
  msw($i$,$v$)},  becomes  exclusively true  (others  false) with  probability
$\theta_{i,v}$ ($v \in  V_i$) where $ \sum_{v \in V_i}  \theta_{i,v} = 1$.  In
other words, ${\tt msw}(i,\cdot)$ stands  for a discrete random variable $X_i$
taking $v$  with probability $\theta_{i,v}$ ($v  \in V_i$).  In  this sense we
identify ${\tt msw}(i,\cdot)$ with $X_i$ and its distribution.

The $\theta_{i,v}$'s are called  {\em parameter\/}s associated with $i$.  They
are  directly  specified  by  the  user  or  learned  from  data.   We  define
$\pmsw(\cdot \mid \btheta)$  as an infinite product of  such distributions for
{\tt  msw}s  where $\btheta$  stands  for the  set  of  all parameters.   Then
$\pmsw(\cdot \mid  \btheta)$ is  uniquely extended by  way of the  least model
semantics  for  logic  programs  to a  $\sigma$-additive  probability  measure
$\pdb(\cdot\mid  \btheta)$  over possible  Herbrand  interpretations of  $\db$
which  we consider as  the denotation  of $\db$({\em  distribution semantics})
\cite{Sato95,Sato01g}.  In the following we omit $\btheta$ when the context is
clear for the sake of brevity.

Let $G$ be  a non-{\tt msw} atom which is  ground.  $\pdb(G)$, the probability
of $G$,  can be naively  computed as follows.   First reduce the  top-goal $G$
using Prolog's exhaustive top-down proof search to an equivalent propositional
DNF formula $\expl_0(G) = \epsilon_1 \vee\cdots\vee \epsilon_k$\footnote{
The equivalence means  that $G$ and $\expl_0(G)$ denote  the same Boolean random
variable in view of the distribution semantics of PRISM.
} \footnote{
When convenient, we treat  $\expl_0(G)$ as a bag $\{\epsilon_1,\ldots,\epsilon_k
\}$ of explanations.
} where  $\epsilon_i$ ($1 \leq i  \leq k$) is a  conjunction $\mbox{\tt msw}_1
\wedge\cdots\wedge \mbox{\tt  msw}_n$ of {\tt msw} atoms  such that $\mbox{\tt
  msw}_1 \wedge\cdots\wedge \mbox{\tt msw}_n,\db \vdash G$.  Each $\epsilon_i$
is called an {\em explanation for G}. Then assuming
\begin{description}
\item[\mbox{[Independence condition]}]
     {\tt msw} atoms in an explanation are independent:\\
      $\pdb(\mswz \wedge \mswz') = \pdb(\mswz)\pdb(\mswz')$
\item[\mbox{[Exclusiveness condition]}]  Explanations are exclusive:\\
      $\pdb(\epsilon_i \wedge \epsilon_j) =0$ if $i \neq j$
\end{description}
we  compute  $\pdb(G)$  as
\begin{eqnarray*}
\pdb(G) &  = &  \pdb(\epsilon_1)   +  \cdots   + \pdb(\epsilon_k) \\
\pdb(\epsilon_i) &  = & \pdb(\mbox{\tt  msw}_1)  \cdots
                   \pdb(\mbox{\tt msw}_n)
  \hspace{1em}\mbox{for}\hspace{1em}
             \epsilon_i = \mbox{\tt msw}_1 \wedge\cdots\wedge \mbox{\tt msw}_n.
\end{eqnarray*}
Recall here  that {\tt  msw}s with different  switch names are  independent by
construction of  $\pmsw(\cdot\mid \btheta)$.  We may further  assume that {\tt
  msw} atoms  with the same switch  name are iid  (independent and identically
distributed). That is when {\tt msw($i$,$v$)} and {\tt msw($i$,$v'$)} occur in
a  program,  we consider  they  are  the results  of  sampling  the same  {\tt
msw($i$,$\cdot$)}  twice.  This  is  justified by  hypothetically adding  an
implicit argument, trial-id $t$  \cite{Sato01g}, to {\tt msw($i$,$\cdot$)} and
assume that {\tt msw($i$,$t$,$\cdot$)}s  have a product of joint distributions
just like the  case of {\tt msw/2} which  makes {\tt msw($i$,$t$,$\cdot$)} and
{\tt msw($i$,$t'$,$\cdot$)} ($t  \neq t'$) iid.  So in  what follows we assume
the independence condition is automatically satisfied.

Contrastingly   the   exclusiveness    condition   cannot   be   automatically
satisfied. It  needs to be  satisfied by the  user, for example, by  writing a
program so that  it generates an output solely as  a sequence of probabilistic
choices  made   by  {\tt   msw}  atoms  (modulo   auxiliary  non-probabilistic
computation).  Although  most generative models including BNs,  HMMs and PCFGs
are written this  way, naturally, but there are models  which are unnatural or
difficult to write this way  \cite{DeRaedt07}.  Relating to this, observe that
{\em Viterbi explanation\/}, i.e. the most likely explanation $\epsilon^*$ for
$G$, is  computed similarly  to $\pdb(G)$ just  by replacing sum  with argmax:
$\epsilon^* \ddefined {\rm  argmax}_{\epsilon \in \expl_0(G)} \pdb(\epsilon)$,
and does  not require the exclusiveness  condition to compute  because it only
deals  with the probability  of a  single explanation.   We will  discuss more
about the exclusiveness condition in Section~\ref{sec:exclusivness}.

\subsection{Tabled search, dynamic programming, probability computation  and Viterbi inference} 
\label{sbusec:tsearch}
So  far  our computation  is  naive. Since  there  can  be exponentially  many
explanations,   naive    computation   would   lead    to   exponential   time
computation. PRISM  avoids this  by adopting tabled  search in  the exhaustive
search  for  all explanations  for  the  top-goal  $G$ and  applying  dynamic
programming to  probability computation.  By  tabling, a goal once  called and
proved is stored (tabled) in  memory with answer substitutions and later calls
to  the  goal  return  with  stored answer  substitutions  without  processing
further.  Tabling is important to probability computation because tabled goals
factor  out common sub-conjunctions  in $\expl_0(G)$,  which results  in sharing
probability  computation for  the common  sub-conjunctions,  thereby realizing
dynamic programming  which gives exponentially  faster probability computation
compared to naive computation.

As a  result of exhaustive tabled  search for all explanations  for $G$, PRISM
obtains a  set of propositional  formulas called {\em defining  formula\/}s of
the  form $H  \Leftrightarrow B_1  \vee\ldots\vee B_h$  for every  tabled goal
$H$\footnote{
The top-goal $G$ is a  tabled goal.  Tabled goals except the top-goal are
called ``intermediate goals'' in \citeN{Sato01g}, \citeN{Zhou08}.
} that directly or indirectly calls {\tt msw}s.  We call the heads of defining
formulas {\em defined goal\/}s.  Each $B_i$ ($1 \leq i \leq h$) is recursively
a  conjunction $C_1 \wedge\ldots\wedge  C_m \wedge  \mswz_1 \wedge\ldots\wedge
\mswz_n$ ($0 \leq m,n$) of defined  goals $\{ C_1,\ldots,C_m \}$ and {\tt msw}
atoms $\{ \mswz_1,\ldots,\mswz_n \}$.  We introduce a binary relation $H \succ
C$ over defined goals  such that $H \succ C$ holds if $H$  is the head of some
defining formula and $C$ occurs in the body.  Assuming ``$\succ$'' is acyclic,
we  extend it to  a partial  ordering over  the defined  atoms.  We  denote by
$\expl(G)$ the  whole set  of defining formulas  and call $\expl(G)$  the {\em
  explanation graph\/} for $G$ like the non-tabled case.

Once  $\expl(G)$  is  obtained,  since   defined  goals  are  layered  by  the
``$\succ$'' relation  by our  assumption where the  a defining formula  in the
bottom layer has only {\tt msw}s in the body whose probabilities are known, we
can compute probabilities by sum-product operation\footnote{
The  exclusiveness and  independence conditions  are inherited  from  the naive
case.
} for all defined goals from  the bottom layer upward in a dynamic programming
manner in time  linear in the size of $\expl(G)$, i.e.\  the number of atoms
appearing in  $\expl(G)$.

Compared to  naive computation, dynamic  programming on $\expl(G)$  can reduce
time  complexity   for  probability  computation  from   exponential  time  to
polynomial time.   For example PRISM's probability computation  for HMMs takes
$O(L)$  time for  a given  sequence  with length  $L$ and  coincides with  the
well-known forward-backward algorithm  for HMMs.  Likewise PRISM's probability
computation for PCFGs  takes $O(L^3)$ time for a sentence  with length $L$ and
coincides  with  the  computation  of  inside  probability  for  PCFGs.   More
interestingly,  BP (belief propagation),  one of  the standard  algorithms for
probability   computation  for   BNs,  coincides   with   PRISM's  probability
computation   applied  to   PRISM  programs   that  describe   junction  trees
\cite{Sato07a}.

Viterbi inference,  i.e.\ the computation  of the Viterbi explanation  and its
probability, is similarly  performed on $\expl(G)$ in a  bottom-up manner like
probability  computation stated  above.  The  only difference  is that  we use
argmax  instead  of  sum.  In  what  follows  we  look  into how  the  Viterbi
explanation is computed. We use $\btheta$  for the set of all parameters.  Let
$H$  be a  defined goal  and $H  \Leftrightarrow B_1  \vee\ldots\vee  B_h$ the
defining formula for  $H$ in $\expl(G)$.  Write $B_i  = C_1 \wedge\ldots\wedge
C_m   \wedge    \mbox{\tt   msw($i_1$,$v_1$)}   \wedge\ldots\wedge   \mbox{\tt
  msw($i_n$,$v_n$)}$  ($1 \leq  i \leq  h$) and  suppose recursively  that the
Viterbi explanation  $\epsilon^*_{C_j}$ ($1 \leq  j \leq m$) has  already been
calculated  for  each  defined goal  in  $C_j$  in  $B_i$.  Then  the  Viterbi
explanation   $\epsilon^*_{B_i}$  for  $B_i$   and  the   Viterbi  explanation
$\epsilon^*_{H}$ for $H$ are respectively computed by
\begin{eqnarray*}
\epsilon^*_{B_i} & = &
     \epsilon^*_{C_1} \wedge\cdots\wedge \epsilon^*_{C_m} \wedge 
        \mbox{\tt   msw($i_1$,$v_1$)} \wedge\ldots\wedge \mbox{\tt   msw($i_n$,$v_n$)} \\
\epsilon^*_{H}  & = & {\rm argmax}_{B_i} \pdb(\epsilon^*_{B_i} \mid \btheta) \\
  &   & \hspace{2em}\mbox{where}\;\;
           \pdb(\epsilon^*_{B_i} \mid \btheta) =  \pdb(\epsilon^*_{C_1})\cdots \pdb(\epsilon^*_{C_m})
                                  \theta_{i_1,v_1} \cdots \theta_{i_n,v_n}.
\end{eqnarray*}
Here $\theta_{i_1,v_1}$ is a  parameter associated with {\tt msw($i_1$,$v_1$)}
and  so on.   In this  way the  Viterbi explanation  for the  top-goal  $G$ is
computed in a  bottom-up manner by scanning $\expl(G)$ once  in time linear in
the size of $\expl(G)$, i.e.\  exactly the same time complexity as probability
computation; For example  $O(L)$ for HMMs and $O(L^3)$ for  PCFGs where $L$ is
respectively the length of sequence and that of sentence.

Parameter learning in  PRISM, be it EM, MAP, VB or  VT(explained next), is based
on computation by dynamic programming  on $\expl(G)$.  For example EM in PRISM
computes   generalized   inside    probabilities   and   generalized   outside
probabilities for  defined goals in  $\expl(G)$ using dynamic  programming and
calculates expectations of the number of  occurrences of {\tt msw} atoms in an
SLD proof for  the top-goal to update parameters  in each iteration, similarly
to  the Inside-Outside  algorithm for  PCFGs \cite{Sato01g}.   MAP  (maximum a
posteriori) estimation and VB (variational Bayes) inference are also performed
similarly \cite{Sato09a,Sato08a}.

\section{Viterbi training and PRISM} 
\label{sec:vt-prism}
In this section we adapt VT to the distribution semantics of PRISM and derives
the VT algorithm for PRISM.

\subsection{Viterbi training} 
\label{subsec:vt}
Here we explain the basic  idea of VT without assuming specific distributions.
Let $x$ be hidden variables, $y$  observed ones and $p(x,y \mid \theta)$ their
joint  distribution  with parameters  $\theta$.  We  assume  $x$ and  $y$  are
discrete.  MLE estimates parameters $\theta$  from $y$ as the maximizer of the
(log) likelihood function $L_{EM}(y \mid \theta)$:
\begin{eqnarray*}
L_{EM}(y \mid \theta) & \ddefined & \log\; \sum_{x}\; p(x,y \mid \theta).
\end{eqnarray*}
In  the  case  of MAP  (maximum  a  posteriori)  estimation,  we add  a  prior
distribution  $p(\theta)$  and  use  $L_{MAP}(y  \mid  \theta)$  below  as  an
objective function:
\begin{eqnarray*}
L_{MAP}(y \mid \theta)
  &  \ddefined & \log\; \sum_{x}\; p(x,y \mid \theta)p(\theta).
\end{eqnarray*}

What VT  does is  similar to  MLE and MAP  but it  uses a  different objective
function $L_{VT}(y \mid \theta)$ defined as
\begin{eqnarray*}
L_{VT}(y \mid \theta)
  &  \ddefined  & \log\; {\rm max}_{x}\; p(x,y \mid \theta)p(\theta).
\end{eqnarray*}
VT  estimates  parameters as  the  maximizer  of  $L_{VT}(y \mid  \theta)$  by
coordinate  ascent  that  alternates  the  maximization of  $\log  p(x,y  \mid
\theta)$  w.r.t.\  $x$ and  the  maximization  of  $\log p(x,y  \mid  \theta)$
w.r.t.\ $\theta$:
\begin{eqnarray}
x^{ (n) }       & = & {\rm argmax}_{ x  }\; \log\; p(x,y \mid \theta^{ (n) })
                          \label{eq:max-hidden}  \\
\theta^{ (n+1) } & = & {\rm argmax}_{ \theta }\; \log\; p(x^{ (n) },y \mid \theta)
                          p(\theta)
                          \label{eq:max-param}
\end{eqnarray}

Starting with appropriate initial parameters $\theta^{ (0) }$, VT iterates the
above two steps  and terminates when $x^{  (n+1) } = x^{ (n)  }$ holds (recall
that  random variables  $x$ and  $y$ are  discrete).  Proving  the convergence
property of VT is straightforward.

\begin{eqnarray*}
L_{VT}(y \mid \theta^{ (n+1) })
  &   =  & \log p(x^{ (n+1) },y \mid \theta^{ (n+1) })p(\theta^{ (n+1) }) \\
  & \geq & \log p(x^{ (n) },y \mid \theta^{ (n+1) })p(\theta^{ (n+1) }) \\
  & \geq & \log p(x^{ (n) },y \mid \theta^{ (n) })p(\theta^{ (n) }) \\
  &   =  & L_{VT}(y \mid \theta^{ (n) })
\end{eqnarray*}
So $L_{VT}(y \mid \theta^{ (n) }) \leq L_{VT}(y \mid \theta^{ (n+1) }) \leq 0$
for every  $n=0,1,\ldots$ Since $\{ L_{VT}(y  \mid \theta^{ (n) })  \}_n$ is a
monotonically increasing  sequence with  an upper bound,  it converges  as $n$
goes to infinity.

\subsection{VT for PRISM}
\label{subsec:vt-for-prism}
Here we reformulate VT in the context  of PRISM.  Let $\db$ be a PRISM program
with parameters $\btheta$ and $\pdb(\cdot \mid \btheta)$ a probability measure
defined  by $\db$.  Also  let $\bG  = G_1,\ldots,G_T$  be observed  goals, and
${\rm expl}(G_t)$ ($1 \leq t \leq T$) the set of all explanations $\epsilon_t$
for  $G_t$ such  that $\epsilon_t,  \db \vdash  G_t$.  $\bG  = G_1,\ldots,G_T$
corresponds  to observed variables  $y$ and  $\epsilon_1,\ldots,\epsilon_T$ to
hidden  variables  $x$  in  $p(x,y  \mid \theta)$  respectively  in  equations
(\ref{eq:max-hidden}) and (\ref{eq:max-param}) in Subsection~\ref{subsec:vt}.

Let {\tt  msw($i$,$\cdot$)} be the set  of {\tt msw} atoms  for a multi-valued
random  switch $i$ as  before that  represents a  probabilistic choice  from a
finite set  $V_i$ of  possible outcomes such  that {\tt msw($i$,$v$)}  ($v \in
V_i$) becomes exclusively true with probability $\theta_{i,v}$\footnote{
In PRISM, $V_i$ is declared by {\tt values/2-3} predicate.
}.
Since $\sum_{v \in V_i} \theta_{i,v} =1$ holds, $\btheta_i $ is a point in the
probability simplex.
We put $\btheta_i = \{ \theta_{i,v}  \}_{v \in V_i} $ and $\btheta = \bigcup_i
\btheta_i$ where $i$ ranges over possible switch names.

We  introduce   as  a   prior  distribution  Dirichlet   distribution  $P_{\rm
  Dir}(\btheta_i) \propto \prod_{v \in  V_i} \theta_{i,v}^{ \alpha_{i,v} - 1 }
$ with  hyper parameters $\{ \alpha_{i,v}  \}_{ v \in V_i}$  over $\btheta_i $
and their product distribution  $P_{\rm Dir}(\btheta) \ddefined \prod_i P_{\rm
  Dir}(\btheta_i)$.    In  the   following,   to  avoid   the  difficulty   of
zero-probability  encountered in  parameter  learning, we  assume {\em  pseudo
  count\/}   $\delta_{i,v}  \ddefined   \alpha_{i,v}  -   1  >   0$   and  use
$\delta_{i,v}$ in place of $\alpha_{i,v}$.

Finally recall the {\em Viterbi explanation\/}  $\epsilon_t^*$ for a goal $G_t$
is a most probable explanation for $G_t$ given by
\begin{eqnarray}
\epsilon_t^* 
   & = & {\rm argmax}_{\epsilon_t \in  {\rm expl}(G_t) } \pdb(\epsilon_t \mid \btheta).
              \label{eq:viterbi}
\end{eqnarray}

By     substituting     $\bG     =     G_1,\ldots,G_T$     for     $y$     and
$\epsilon_1,\ldots,\epsilon_T$  for $x$  in the  definition of  $L_{VT}(y \mid
\theta)$, the objective function $L_{VT}(\bG \mid \btheta)$ for VT in PRISM is
now computed as follows.
\begin{eqnarray}
\lefteqn{ L_{VT}(\bG \mid \btheta) } \nonumber \\
  & = & \log\; {\rm max}_{ \epsilon_1 \in {\rm expl}(G_1),\ldots,\epsilon_T \in {\rm expl}(G_T)}
                   \prod_{ t=1 }^T \; \pdb(\epsilon_t, G_t \mid \btheta) P_{\rm Dir}(\btheta)  \nonumber \\
  & = & \log\; \prod_{ t=1 }^T \;  {\rm max}_{ \epsilon_t \in {\rm expl}(G_t)} \pdb(\epsilon_t \mid \btheta)P_{\rm Dir}(\btheta) 
                 \nonumber \\
  & = & \log\; \prod_{ t=1 }^T \; \pdb(\epsilon_t^* \mid \btheta) P_{\rm Dir}(\btheta) \nonumber \\
  & = & \log\; \prod_{i,v} \theta_{i,v}^{ \sum_{t=1}^T  \sigma_{i,v}(\epsilon_t^*) + \delta_{i,v} } \nonumber  \\
  & = & \sum_{i,v} \left( \sum_{t=1}^T  \sigma_{i,v}(\epsilon_t^*) + \delta_{i,v} \right) \log \theta_{i,v}
              \label{eq:objective-LVT}
\end{eqnarray}
where ``$i,v$'' ranges over those such that {\tt msw($i$,$v$)} appears in some
$\epsilon_t^*$   and  $\sigma_{i,v}(\epsilon_t^*)$  is   the  count   of  {\tt
  msw($i$,$v$)} in $\epsilon_t^*$.

Likewise   by    substituting   $\bG    =   G_1,\ldots,G_T$   for    $y$   and
$\epsilon_1,\ldots,\epsilon_T$ for $x$  in equations (\ref{eq:max-hidden}) and
(\ref{eq:max-param}) respectively and  using the definition of $\epsilon_t^*$,
we   obtain  the   VT   algorithm  for   PRISM   which  alternately   executes
(\ref{eq:max-hidden-prism})  and  (\ref{eq:max-param-prism}) where  $\btheta^{
  (n) }$ stands for the set of parameters $\{ \theta_{i,v}^{ (n) } \}$ at step
$n$.

\begin{eqnarray}
\epsilon_t^{ *(n) }  &  =  &
          {\rm argmax}_{ \epsilon_t \in  {\rm expl}(G_t) }\;
                  \pdb( \epsilon_t \mid \btheta^{ (n) })  \;\;\; (1 \leq t \leq T)
                          \label{eq:max-hidden-prism}  \\
\theta_{i,v}^{ (n+1) } & \propto &
           \sum_{t=1}^T  \sigma_{i,v}(\epsilon_t^{ *(n) }) + \delta_{i,v}
                          \label{eq:max-param-prism}
\end{eqnarray}
Here  (\ref{eq:max-hidden-prism})  corresponds  to  (\ref{eq:max-hidden})  and
(\ref{eq:max-param-prism})   to   (\ref{eq:max-param})  respectively.

Using (\ref{eq:max-hidden-prism}) and  (\ref{eq:max-param-prism}), VT in PRISM
is  performed as  follows.  Given  observed goals  $\bG = G_1,\ldots,G_T$, we
first perform tabled search for  all explanations to build explanations graphs
$\expl(G_t)$ for each $t$ ($1 \leq t \leq T$).  Then starting from the initial
parameters  $\btheta^{  (0)  }$,  we  repeat  (\ref{eq:max-hidden-prism})  and
(\ref{eq:max-param-prism})    alternately   while   computing    the   Viterbi
explanations $\epsilon_t^{  *(n) }$ in  (\ref{eq:max-hidden-prism}) by dynamic
programming over  $\expl(G_t)$ as explained  in Section~\ref{sec:review-prism}
until $\epsilon_t^{ *(n+1) } = \epsilon_t^{ *(n) }$ holds for all $t$ ($1 \leq
t \leq T$).  $\{ \theta_{i,v}^{ (n+1) } \}$ are then learned parameters.\\

Having  derived the  VT algorithm  for  PRISM, we  examine the  effect of  the
termination condition $\epsilon_t^{ *(n+1) } = \epsilon_t^{ *(n) }$ ($1 \leq t
\leq T$)  on the  convergence of VT.   As we remarked  in Section~\ref{intro},
this  condition  means VT  terminates  as  soon  as the  Viterbi  explanations
converge, i.e.\  there is no change  of the Viterbi  explanations between step
$n$ and step $n+1$ whereas EM always runs until convergence of parameters.  As
a  result since  a small  change  of parameters  does not  affect the  Viterbi
explanation but keeps EM running, VT  tends to converge in much less number of
iterations than EM.

To empirically  check this, we  conducted parameter learning  of probabilistic
grammars  by  VT  and  by  EM  using  PRISM  and  compared  their  convergence
behavior\footnote{
All experiments in this  paper are done on a single machine  with Core i7 Quad
2.67GHz$\times$2  CPU and 72GB RAM running OpenSUSE 11.2, using PRISM2.1.
}.   We used  two probabilistic  grammars, a  PCFG and  a  PLCG (probabilistic
left-corner  grammar) for  the ATR  corpus \cite{Uratani94}(their  details are
described in the next section),  and measured the average number of iterations
and learning time\footnote{
We used a built-in predicate {\tt prism\_statistics(em\_time,$x$)} to measure
learning time which returns in $x$ time used by the learning algorithm.
}   required  for  convergence   over  ten   runs.   Table~\ref{table:ave-itr}
summarizes the results with standard deviations in parentheses.

\begin{table}[h]
\begin{center}
\caption{ Average number of iterations and learning time for convergence }
\label{table:ave-itr}
\begin{tabular}{ccccc} \hline
\rule{0pt}{12pt}
         & \multicolumn{2}{c}{Iterations}
         & \multicolumn{2}{c}{Learning time(sec)} \\ \cline{2-5}
\rule{0pt}{12pt}  \\ [-6pt]
         &   VT         & EM           & VT         &  EM     \\ \hline\hline
PCFG     &   8.10(2.28) & 123.6(3.23)  & 0.45(0.11) &  6.29(0.16)   \\ \hline
PLCG     &  15.80(4.73) & 144.2(43.51) & 1.55(0.36) & 11.686(3.64)    \\ \hline
\end{tabular}
\end{center}
\end{table}

Looking  at  the table,  we  see  that VT  required  only  a  small number  of
iterations  to  converge  compared to  EM;  the  ratio  of average  number  of
iterations of VT to EM is 1:15.2  w.r.t.\ the PCFG and 1:8.3 w.r.t.\ the PLCG.
We also note that the ratio of average learning time\footnote{
Learning  time displayed  by the  PRISM system  after {\tt  learn}  is ``total
learning time'' which includes search time for explanations and other overhead
time such as copying {\tt msw}s  in the memory, in addition to actual learning
time reported by {\tt prism\_statistics(em\_time,$x$)}.  Since such extra-time
accounts for  a large percent of total  learning time, it can  happen that the
difference  in  total  learning  time  between  EM  and  VT  is  smaller  than
Table~\ref{table:ave-itr}.
} is  similar  to  that  of  iterations,  1:13.8 w.r.t.\  the  PCFG  and  1:7.4
w.r.t.\ the PLCG  respectively.  It therefore seems natural  to conclude that
VT learns  parameters with  much less number  of iterations and  thereby much
faster than EM\footnote{
In the  table, the  difference of  VT and EM  in the  number of  iterations is
statistically  significant for  both grammars  by unpaired  t-test at  the 5\%
significance level  with the Bonferroni correction.  This  applies to learning
time as well.
}.

Since VT is  a local maximizer, it is sensitive to  the initial condition like
EM.  So we need to  carefully choose $\btheta^{ (0) }$.  Uniform distributions
for $\btheta^{ (0) }$ \cite{Spitkovsky10}  and $\epsilon_t^{ *(0) }$($1 \leq t
\leq T$)  \cite{Cohen10} are  possible choices.  In  practice, we  further add
random  restart to  alleviate the  sensitivity  problem.  For  example in  the
experiments in the next section,  we repeated parameter learning 50 times with
random restart  for each  learning and selected  the parameter set  giving the
largest value of the objective function $L_{VT}(\bG \mid \btheta)$ computed by
(\ref{eq:objective-LVT}).

\section{Learning experiments with probabilistic grammars} 
\label{sec:grammar}

In this  section we apply VT  to parsing tasks in  natural language processing
where observable variables are sentences and hidden variables are parse trees.
We predict parse trees for  given sentences using probabilistic grammars (PCFG
and  PLCG)  whose  parameters  are  learned  by VT  and  compare  the  parsing
performance with each of EM, MAP and VB\footnote{
We assume the reader is familiar with the basics of parsing theory.
}.

\subsection{VT for PCFGs} 
Prior to  describing the parameter learning  experiment with a PCFG  by VT, we
briefly review how to write PCFGs  in PRISM.  In PCFGs, sentence derivation is
carried  out probabilistically.   When there  are $k$  PCFG  rules $\theta_1:A
\rightarrow \beta_1,\ldots, \theta_k:A  \rightarrow \beta_k$ for a nonterminal
$A$  with probabilities  $\theta_1,\ldots,\theta_k$ ($\theta_1+\cdots+\theta_k
=1$),  $A$  is  expanded  by  $A  \rightarrow  \beta_i$  into  $\beta_i$  with
probability $\theta_i$.  The probability of a parse tree $\tau$ is the product
of probabilities  associated with occurrences of  CFG rules in  $\tau$ and the
probability of a  sentence is the sum of probabilities of  parse trees for the
sentence.

Writing PCFG programs is easy  in PRISM.  Fig.~\ref{prog:pcfg}  is a PRISM
program  for   a  PCFG  \{ {\tt   0.4:S$\rightarrow$S  S,  0.3:S$\rightarrow$a,
  0.3:S$\rightarrow$b} \}. In    general,     PCFG    rules     such    as
\{ $\theta_1:A\rightarrow\beta_1,\ldots,  \theta_k:A\rightarrow\beta_k$ \} are
encoded by {\tt values/3} declaration as
\begin{center}
{\tt values('A',[$\beta_1,\ldots,\beta_k$],[$\theta_1,\ldots,\theta_k$])}
\end{center}
where $\beta_i$ ($1 \leq i \leq k$) is a Prolog list of terminals and nonterminals.

\begin{figure}[b]
\rule{1.0\hsize}{0.8pt}
\begin{verbatim}
values('S',[['S','S'],[a],[b]],[0.4,0.3,0.3]).

pcfg(L):- pcfg(['S'],L,[]).
pcfg([A|R],L0,L2):-
  ( get_values(A,_) ->  % msw(A,_) exists, so
      msw(A,RHS),       % A is a nonterminal
      pcfg(RHS,L0,L1)
  ; L0=[A|L1] ),
  pcfg(R,L1,L2).
pcfg([],L,L).
\end{verbatim}
\rule{1.0\hsize}{0.8pt}
\caption{ A PCFG program }
\label{prog:pcfg}
\end{figure}

We  wrote a  PCFG  program as  shown  in Fig.~\ref{prog:pcfg}  for the ATR  corpus
\cite{Uratani94} using an associated CFG\footnote{
In  the experiment,  to  speed up  parsing,  we partially  evaluated the  PCFG
program with  individual CFG  rules and used the resulting specialized program.
}.   The corpus  contains labeled  parse trees  for 10,995  Japanese sentences
whose average length  is about 10.  The associated  CFG\footnote{
copy right protected.
} comprises 861 CFG rules (168 non-terminals and 446 terminals) and yields 958
parses/sentence on  average.  We applied  four learning algorithms,  i.e.\ VT,
EM, MAP  and VB \cite{Sato09a} available  in PRISM2.1 to the  PCFG program for
the ATR corpus\footnote{
In PRISM, EM  is a special case  of MAP inference.  We used  random but almost
uniform  initialization   of  parameters  and  set   uniformly  pseudo  counts
$\delta_{i,v}$  to  $\mbox{1.0}^{-9}$   for  EM  and  1.0  for   MAP  and  VT,
respectively.  Similarly  we uniformly set hyper  parameters $\alpha_{i,v}$ to
1.0 for VB.  The number of  candidates for re-ranking in VB \cite{Sato09a} was
set to 5.

In all  cases, we set  the number of  random restart to  50 and used  the best
parameter   set  that  gave   the  largest   value  of   objective  functions,
i.e.\ $L_{\rm EM}$ for EM, $L_{\rm MAP}$ for MAP and $L_{\rm VT}$ for VT.  For
the case  of VB that  learns pseudo  counts, we chose  the best set  of pseudo
counts giving the highest free energy \cite{Sato09a}.
} and compared the performance of VT with other learning methods.\\

We conducted  eight-fold CV (cross validation)  for each algorithm\footnote{
We chose eight-fold CV for parallel execution of learning by our
machine.
} to evaluate the quality of  learned parameters in terms of three performance
metrics  i.e.\ LT(labeled  tree),  BT(bracketed tree)  and 0-CB(zero  crossing
brackets) \cite{Goodman96}.  These metrics are  computed from $T_c$, the set of
parse trees in  a test corpus which are considered correct  and $T_g$, the set
of  parse trees  predicted  for sentences  in  the test  corpus  by a  parsing
algorithm.  LT is defined as $|T_c \cap T_g|/N$ where $|S|$ denotes the number
of elements  in a set $S$ and  $N=|T_g|=|T_c|$.  It is the  ratio of correctly
predicted  labeled parse trees  to the  total number  of labeled  parse trees.
Compared to LT, BT is a  less strict metric that ignores nonterminals in parse
trees.   Let  $T'_g$  be the  set  of  unlabeled  trees obtained  by  removing
nonterminals from $T_g$ which  coincide with the corresponding unlabeled trees
in $T_c$.  Then BT is defined as $|T'_g|/N$.  Finally 0-CB is the least strict
metric in  the three  metrics.  We say  brackets $(w_i,\ldots,w_j)$ in  a tree
$\tau$ is inconsistent with another  tree $\tau'$ if $\tau'$ contains brackets
$(w_{s},\ldots,w_{t})$ such that $s < i \leq t< j$ or $i<s \leq j<t$. Otherwise
they are  consistent with $\tau'$.  Let $T''_g$  be the set of  trees in $T_g$
which have  no inconsistent  brackets with the  corresponding trees  in $T_c$.
Then 0-CB is given by $|T''_g|/N$.\\

To  perform cross  validation, the  entire  corpus is  partitioned into  eight
sections.  In each fold, one section is used as a test corpus and sentences in
the remaining sections are used as training data.  For each of EM, MAP, VT and
VB, parameters (or pseudo counts) are learned from the training data.  A parse
tree is predicted, i.e.\ the Viterbi explanation is computed for each sentence
in  the  test corpus  using  learned parameters  or  using  the approximate  a
posterior distribution  learned by  VB.  The predicted  trees are  compared to
answers, i.e.\ the labeled trees in the test corpus to compute LT, BT and 0-CB
respectively.  The  final performance figures are calculated  as averages over
eight  folds and  summarized in  Table~\ref{table:parsing-pcfg}  with standard
deviations in parentheses.

\begin{table}[h]
\begin{center}
\caption{ Parsing performance by PCFG }
\label{table:parsing-pcfg}
\begin{tabular}{ccccc} \hline
\rule{0pt}{12pt}
           &    \multicolumn{4}{c}{Learning method} \\ \cline{2-5}
\rule{0pt}{12pt}  \\ [-6pt]
Metric   &  VT               & EM          & MAP         & VB         \\ \hline\hline
LT(\%)   & {\bf 74.69}(0.87) & 70.02(0.88) & 70.31(1.13) & 72.13(1.10)       \\ \hline
BT(\%)   & {\bf 77.87}(0.84) & 73.10(1.01) & 73.45(1.20) & 75.46(1.13)       \\ \hline
0-CB(\%) & 83.78(0.92)       & 84.44(0.89) & 84.89(0.84) & {\bf 87.08}(0.87) \\ \hline
\end{tabular}
\end{center}
\end{table}

We statistically analyzed  the parsing  performance by  Dunnett's test\footnote{
We  used Dunnett's  test for  multiple  comparisons of  means with  VT as  the
control  to avoid  inflating the  significance  level.  Figures  in bold  face
indicate best performance.
}. The result is that VT outperformed all of EM, MAP and VB in terms of LT and
BT at the 5\% level of significance but  did not so in terms of 0-CB.  This is
understandable if we assume that there are many parse trees that can give high
scores  in  terms of  less  restrictive  metrics such  as  0-CB  but since  VT
concentrates  probability mass  on a  single tree,  those promising  trees are
allocated  little probability  mass by  VT,  which results  in relatively  low
performance of VT in terms of 0-CB.

So far we examined parsing  performance by parameters obtained from incomplete
data (sentences  in the corpus).   We also examined parsing  performance using
8-fold  CV by  parameters  learned  from complete  data,  i.e.\ by  parameters
obtained by  counting occurrences of CFG  rules in the corpus.   The result is
LT:79.06\%(1.25), BT:85.28\%(0.69),  0-CB:95.37\%(0.26)(figures in parentheses
are standard deviations).   These figures are considered as  the best possible
performance.  We  notice the gap  in parsing performance between  the complete
data  case  and  the  incomplete  data  case tends  to  become  wider  as  the
performance metric gets less restrictive in the order of LT, BT and 0-CB.

Another thing to note  is that the objective functions for EM,  MAP and VB are
similar  in the  sense that  they  all sum  out hidden  variables whereas  the
objective   function  for   VT  retains   them.   This   fact   together  with
Fig.~\ref{table:parsing-pcfg}  seems to  suggest that  parsing  performance is
more affected by the difference  among objective functions than the difference
among learning methods.

\subsection{ VT for PLCGs } 

\begin{figure}[t]
\rule{1.0\hsize}{0.8pt}
{\footnotesize
\begin{verbatim}
values(lc('S','S'),[rule('S',['S','S'])]).
values(lc('S',a),[rule('S',[a])]).
values(lc('S',b),[rule('S',[b])]).
values(first('S'),[a,b]).
values(att('S'),[att,pro]).

plcg(L):- g_call(['S'],L,[]).

g_call([],L,L).
g_call([G|R],[Wd|L],L2):-
   ( G = Wd -> L1 = L     % shift operation
   ; msw(first(G),Wd),lc_call(G,Wd,L,L1) ),
   g_call(R,L1,L2).

lc_call(G,B,L,L2):-       % B-tree is completed   
   msw(lc(G,B),rule(A,[B|RHS2])),
   ( G = A -> true ; values(lc(G,A),_) ),
   g_call(RHS2,L,L1),     % complete A-tree
   ( G = A -> att_or_pro(A,Op), 
     ( Op = att -> L2 = L1 ; lc_call(G,A,L1,L2) )
   ; lc_call(G,A,L1,L2) ).

att_or_pro(A,Op):-
   ( values(lc(A,A),_) -> msw(att(A),Op) ; Op=att ).
\end{verbatim}
}
\rule{1.0\hsize}{0.8pt}
\caption{ A PLCG program }
\label{prog:plcg}
\end{figure}

PCFGs   assume  top-down  parsing.    Contrastingly  there   is  a   class  of
probabilistic  grammars  based on  bottom-up  parsing  for  CFGs called  PLCGs
(probabilistic    left-corner   grammars)   \cite{Manning97,Roark99,Uytsel01}.
Although they use the same set  of CFG rules as PCFGs but attach probabilities
not  to  expansion of  nonterminals  but  to  three elementary  operations  in
bottom-up parsing, i.e.\ shift, attach and  project. As a result they define a
different class of distributions from PCFGs.  In this subsection we conduct an
experiment for parameter learning of a PLCG by VT.

The objective of this  subsection is two fold.  One is to  apply VT to a PLCG,
which seems not attempted before as far as we know, and to examine the parsing
performance.  The other is to  empirically demonstrate the universality of our
approach  to   VT  that  subsumes  differences  in   probabilistic  models  as
differences in  explanation graphs  and applies a  single VT algorithm  to the
latter.

Programs   for   PLCGs   look   very   different   from   those   for   PCFGs.
Fig.~\ref{prog:plcg} is  a PLCG program  which is a  dual version of  the PCFG
program  in   Fig.~\ref{prog:pcfg}  with   the  same  underlying   CFG  \{{\tt
  S$\rightarrow$S   S,  S$\rightarrow$a,  S$\rightarrow$b}\}.    It  generates
sentences using  the first set of  {\tt 'S'} and the  left-corner relation for
this  CFG ({\tt values/2}  there only  declares the  space of  outcomes).  The
program works as follows. Suppose nonterminals  {\tt G} and {\tt B} are in the
left-corner  relation and  {\tt G}  is  waiting for  a {\tt  B}-tree, i.e.\  a
subtree with  the root  node labeled {\tt  B}, to  be completed.  When  a {\tt
  B}-tree is  completed, the program  probabilistically chooses a CFG  rule of
the form ${\tt  A} \rightarrow {\tt B}\beta$ to further  grow the {\tt B}-tree
using this  rule.  Upon the completion  of the {\tt  A}-tree and if {\tt  G} =
{\tt A}, the  attach operation or the projection  is probabilistically chosen.
By  replacing  {\tt  values}   declarations  appropriately,  this  program  is
applicable to any PLCG.

\begin{table}[t]
\begin{center}
\caption{ Parsing performance by PLCG }
\label{table:parsing-plcg}
\begin{tabular}{ccccc} \hline
\rule{0pt}{12pt}
          &    \multicolumn{4}{c}{Learning method} \\ \cline{2-5}
\rule{0pt}{12pt}  \\ [-6pt]
Metric   &  VT               & EM          & MAP         & VB           \\ \hline\hline
LT(\%)   & {\bf 76.26}(0.96) & 71.81(0.91) & 71.17(0.93) & 71.15(0.90)  \\ \hline
BT(\%)   & {\bf 78.86}(0.70) & 75.17(1.15) & 74.28(1.12) & 74.28(1.00)  \\ \hline
0-CB(\%) & {\bf 87.45}(1.00) & 86.49(0.97) & 86.03(0.67) & 86.04(0.71)  \\ \hline
\end{tabular}
\end{center}
\end{table}

We have  developed a PLCG  program similarly to  the PCFG program for  the ATR
corpus and applied VT, EM, MAP and  VB to it to learn parameters.  We measured
parsing  performance by  learned parameters  in terms  of LT,  BT and  0-CB by
eight-fold   CV   for   each   of   VT,   EM,  MAP   and   VB   and   obtained
Table~\ref{table:parsing-plcg}  (standard   deviations  in  parentheses).   We
compared the parsing  performance of VT with EM, MAP and  VB by Dunnett's test
at  the 5\%  level of  significance  similarly to  the PCFG  case.  This  time
however VT outperformed all of EM, MAP and VB by all metrics, i.e.\ LT, BT and
0-CB.

\section{Applying VT to classification tasks}
\label{sec:nbh}
In the previous  section, we conducted learning experiments with  a PCFG and a
PLCG  in which  the prediction  target was  parse trees  that coincide  with a
hidden variable  in a probabilistic  model.  In this  section, we deal  with a
different situation where a prediction  target differs from a hidden variable.
We apply  VT to classification tasks  using an {\em NBH\/}(naive  Bayes with a
hidden  variable) model whose  hidden variable  is summed  out and  instead an
observable variable, class label, is predicted for the given data.

\begin{figure}[ht]
\centerline{\includegraphics[scale=0.55]{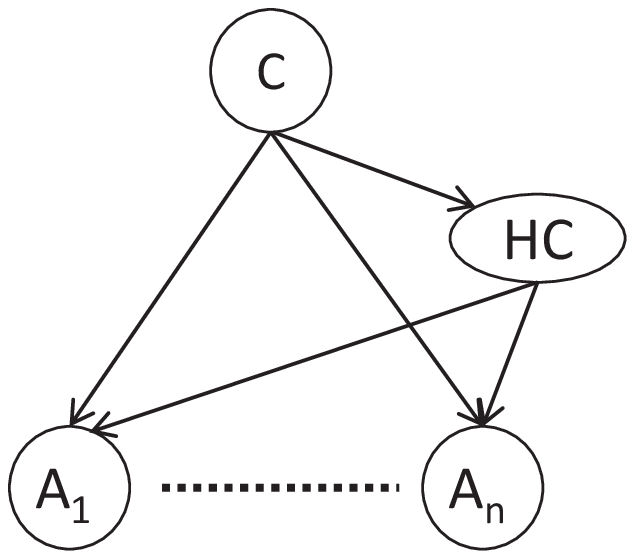}}
\caption{ A Bayesian network for NBH model } \label{fig:NBH1}
\end{figure}

Before  explaining  classification  tasks,  we  review  NBH  for  completeness
\cite{Sato11b}. NBH  is an extension of  NB (naive Bayes) with  a hidden class
variable  $HC$ as  illustrated  in Fig.~\ref{fig:NBH1}.   It  defines a  joint
distribution
\begin{eqnarray*}
P(A_1,\ldots,A_n,HC,C \mid \btheta) 
  & = &  \prod_{j=1}^n P(A_j \!\mid\! HC,C, \btheta)
            P(HC \!\mid\! C, \btheta)P(C \mid  \btheta)
\end{eqnarray*}
where  \btheta\/  is model  parameters,  the  $A_j$'s  attributes of  observed
data\footnote{
We  interchangeably  use the  attributes  $A_1,\ldots,A_n$  as  data when  the
context is clear.
}, $C$ a class  and $HC$ a hidden class.  It is  easily seen from the equation
(\ref{eq:mixture}) below that NBH represents  the data distribution in a class
$C$ as a mixture of data distributions indexed by $HC$.
\begin{eqnarray}
P(A_1,\ldots,A_n \mid C, \btheta) 
  & =  & \sum_{HC} 
         P(A_1,\ldots,A_n  \!\mid\!  HC,C, \btheta)P(HC \!\mid\! C, \btheta) \nonumber \\
  & = & \sum_{HC} 
          \prod_{j=1}^n P(A_j \!\mid\! HC,C, \btheta)P(HC \!\mid\! C, \btheta)
                  \label{eq:mixture}
\end{eqnarray}
The role  of $HC$ is  to cluster data  in a class  $C$ so that  a distribution
$P(A_1,\ldots,A_n \!\mid\!   HC,C, \btheta)$  in each cluster  $HC$ satisfies
the  independent condition  $P(A_1,\ldots,A_n  \!\mid\!  HC,C, \btheta)=  \prod_{j=1}^n
P(A_j \!\mid\!  HC,C, \btheta)$ imposed on NB  as much as possible.  NBH was introduced
in \cite{Sato11b} as  a simple substitute for more  complicated variants of NB
such as TAN \cite{Friedman97},  AODE \cite{Webb05}, BNC \cite{Castillo05}, FBC
\cite{Su06} and HBN \cite{Jiang09}.

Given data $A_1,\ldots,A_n$, we classify $A_1,\ldots,A_n$ as a class $C^*$ by
\begin{eqnarray}
C^* & = & {\rm argmax}_C P(C \mid A_1,\ldots,A_n,\btheta) \nonumber \\
    & = & {\rm argmax}_C \sum_{HC}
       P(C \mid HC,A_1,\ldots,A_n,\btheta)P(HC \mid A_1,\ldots,A_n,\btheta).
                  \label{eq:expert-mixture}
\end{eqnarray}
Note here that  the hidden variable, $HC$, is {\em  not\/} a prediction target
unlike probabilistic  grammars. It  is just summed  out.  However we  expect a
sub-classifier $P(C \mid  HC,A_1,\ldots,A_n,\btheta)$ indexed by $HC$ performs
better  than $P(C  \mid  A_1,\ldots,A_n,\btheta)$, the  original  NB, in  each
cluster and so does their mixture (see equation (\ref{eq:expert-mixture})).\\

\begin{figure}[t]
\rule{1.0\hsize}{0.8pt}
\begin{verbatim}
values(class,[democrat,republican]). % class labels are democrat or republican
values(attr(_A,_C,_HC),[y,n]).       % attribute values are y or n

nbayes(C,Vals):-
   msw(class,C),msw(hclass(C),HC),nbh(1,C,HC,Vals).
nbh(J,C,HC,[V|Vals]):-
   ( V == '?' -> msw(attr(J,C,HC),_)  % '?' indicates missing value
   ; msw(attr(J,C,HC),V) ),
   J1 is J+1,
   nbh(J1,C,HC,Vals).
nbh(_,_,_,[]).
\end{verbatim}
\rule{1.0\hsize}{0.8pt}
\caption{ An NBH program }
\label{prog:nbh}
\end{figure}
To evaluate the quality of  parameters learned by VT for classification tasks,
we conducted a  learning experiment with NBH using ten data  sets from the UCI
Machine       Learning          Repository\cite{Frank10}.
In the experiment training data  is given as a set of tuples
$C,A_1,\ldots,A_n$ consisting of a  class $C$ and attributes $A_1,\ldots,A_n$.
Parameters  (or  pseudo  counts) are  learned  by  a  PRISM program  shown  in
Fig.~\ref{prog:nbh}\footnote{
This program is for the  vote data set from the repository.
} which  is  also used  for  predicting  class labels  in  test  data.  A  {\tt
  values/2}  declaration {\verb!values(class,[democrat,republican])!}
in the program tells  PRISM  to introduce   two   {\tt    msw}   atoms,   {\tt
  msw(class,democrat)}  and  {\tt   msw(class,republican)}  that  represent  a
probabilistic choice  between {\tt  democrat} and {\tt  republican} as  a {\tt
  class}, implicitly together  with their parameters $\theta_{class,democrat}$
and    $\theta_{class,republican}$   such    that   $\theta_{class,democrat}$ +
$\theta_{class,republican}$ = 1.   This  program  assumes  that  attributes  are
numbered and missing values in a data set are replaced with {\tt '?'}.

\begin{table*}[b]
\begin{center}
\caption{Accuracy by VT, EM, MAP and VB}
\label{table:acc}
\begin{tabular}{ccccccc} \hline
\rule{0pt}{12pt}
              &        &   NB     & \multicolumn{4}{c}{NBH : Learning method} \\ \cline{4-7}
\rule{0pt}{12pt} \\[-6pt]
Data set      &  Size  &  EM(\%)  &   VT(\%)     &  EM(\%) & MAP(\%) & VB(\%) \\ \hline \hline
nursery       & 12960  &  90.23   &  92.93       &  99.40       &  {\bf 99.65}  & 97.45 \\
mushroom      &  8124  &  99.57   & {\bf 100.00} & {\bf 100.00} & {\bf 100.00}  & 99.99 \\
kr-vs-kp      &  3196  &  87.86   &  88.69       &  91.59       &  {\bf 92.34}  & 88.90 \\
car           &  1728  &  85.86   &  90.97       &  97.67       &  {\bf 97.82}  & 94.68 \\
votes         &   435  &  90.29   &  96.00       &  95.66       &  {\bf 96.51}  & 96.05 \\
dermatology   &   336  &  97.73   &  97.98       &  97.51       &  98.06        & {\bf 98.17} \\
glass         &   214  &  72.82   &  75.86       &  {\bf 76.84} &  76.66        & 76.53 \\
iris          &   150  &  94.40   &  95.07       &  {\bf 95.13} &  95.07        & 95.07 \\
breast-cancer &   150  &  72.52   &  72.52       &  70.07       &  72.76        & {\bf 72.83} \\
zoo           &   101  &  95.07   &  96.55       &  {\bf 97.42} &  96.95        & 96.62 \\ \hline
%
\end{tabular}
\end{center}
\end{table*}

We obtained  the classification accuracy of  NBH for each  combination of data
set, learning  method (VT, EM,  MAP, VB), the  number of clusters $\#HC$  in a
class $C$ (from  2 to 15) and hyper  parameters (\{0.1,1.0\} as $\alpha_{i,v}$
for  VB, and  the same  as  pseudo counts  $\delta_{i,v}$ for  VT,MAP) as  the
average over ten times ten-fold CV\footnote{
We used  ten times ten-fold CV  when possible to have  robust estimates though
computationally expensive \cite{Japkowicz11}.
} except  nursery, mushroom and kr-vs-kp  data sets in which  case ten-fold CV
was  used.   We  similarly  obtained  the classification  accuracy  of  NB  as
baseline\footnote{
We used  the EM algorithm  for parameter learning  of NB as there  are missing
data in some data sets.
}.

Table~\ref{table:acc}  summarizes  classification accuracies  of  NB and  NBH.
Accuracy for NBH in the table  is the best accuracy obtained by varying $\#HC$
and hyper  parameters as we mentioned  for the given learning  method and data
set.  Figures  in bold face indicate  the best accuracy achieved  in each data
set.  The table shows that for most  data sets NBH performed better than NB as
we expected. Actually  the difference in accuracy between NB  and the best one
for NBH is statistically significant by  unpaired t-test at the 5\% level with
the Bonferroni correction\footnote{
As ten data sets are used, the significance level is set to 5\%/10 = 0.5\%.
}  for  all  data  sets  except  dermatology,  iris  and  breast-cancer.   The
superiority of NBH  over NB demonstrated in this  experiment is interpreted as
an effect of clustering in a class by introducing a hidden variable $HC$.

Comparing  the classification  accuracies by  four parameter  learning methods
applied to  NBH, we notice  that VT's performance  is comparable to  the other
three, i.e.\ EM, MAP  and VB except for the case of  nursery, kr-vs-kp and car
data  sets.  For these  data sets  VT's accuracy  is worse  than the  best one
achieved  by  one  of  the  three learning  methods,  which  is  statistically
confirmed  by  unpaired t-test  at  the 5\%  level  of  significance with  the
Bonferroni correction.   So from the  viewpoint of a learning  experiment with
NBH, we cannot  say, regrettably, VT outperformed EM, MAP and  VB for all data
sets.   However, the  result is  understandable if  we recall  that  while the
predication target  in the  experiment is a  class variable $C$,  VT optimizes
parameters not  for $C$ but for the  hidden variable $HC$ which  is summed out
and hence only indirectly affects prediction.

\section{Removing the exclusiveness condition} 
\label{sec:exclusivness}

PRISM assumes the exclusiveness  condition on programs to simplify probability
computation  as  explained  in  Section~\ref{sec:review-prism}.  It  means  we
cannot write a  program clause $H \Leftarrow B \vee  B'$ unless $\pdb(B \wedge
B'  \mid  \btheta)  =  0$  is guaranteed  \cite{Sato01g}.   Although  most  of
generative  probabilistic models  such as  BNs, HMMs  and PCFGs  are naturally
described as PRISM programs satisfying the condition,  removing it certainly
gives us more freedom of probabilistic modeling.  Theoretically it is possible
to  remove  it by  introducing  BDDs  (binary  decision diagrams)  as  ProbLog
\cite{DeRaedt07,Kimmig08}  and  PITA \cite{Riguzzi11}  do,  and their  related
systems,     LeProbLog\cite{Gutmann08},     LFI-ProbLog\cite{Gutmann11}    and
EMBLEM\cite{Bellodi12},   offer  parameter   learning  based   on  probability
computation  by BDDs, though  with different  learning frameworks  from PRISM.
If, however, we are only interested in obtaining the Viterbi explanation after
parameter learning  as we are  in many cases,  VT gives us  a way of  doing it
without  BDDs  even  for  programs  that  do  not  satisfy  the  exclusiveness
condition.  This is because VT does not require the exclusiveness condition to
execute  equations (\ref{eq:max-hidden-prism})  and (\ref{eq:max-param-prism})
that always deal with a single explanation and a single probability.

\begin{figure}[h]
\rule{1.0\hsize}{0.8pt}
{\footnotesize
\begin{verbatim}
values(d_e(1,2),[on,off],[0.9,0.1]). values(d_e(2,3),[on,off],[0.8,0.2]).
values(d_e(3,4),[on,off],[0.6,0.4]). values(d_e(1,6),[on,off],[0.7,0.3]).
values(d_e(2,6),[on,off],[0.5,0.5]). values(d_e(6,5),[on,off],[0.4,0.6]).
values(d_e(5,3),[on,off],[0.7,0.3]). values(d_e(5,4),[on,off],[0.2,0.8]).

d_e(1,2):- msw(d_e(1,2),on). d_e(2,3):- msw(d_e(2,3),on).
d_e(3,4):- msw(d_e(3,4),on). d_e(1,6):- msw(d_e(1,6),on).
d_e(2,6):- msw(d_e(2,6),on). d_e(6,5):- msw(d_e(6,5),on).
d_e(5,3):- msw(d_e(5,3),on). d_e(5,4):- msw(d_e(5,4),on).

path(X,Y) :- path(X,Y,[X]).

path(X,X,_).
path(X,Y,A):- X\==Y, (d_e(X,Z) ; d_e(Z,X)), absent(Z,A), path(Z,Y,[Z|A]).

absent(_,[]).
absent(X,[Y|Z]):- X\==Y, absent(X,Z).
\end{verbatim}
}
\rule{1.0\hsize}{0.8pt}
\caption{ A graph program violating the exclusiveness condition }
\label{prog:problog_graph}
\end{figure}

We  next  give  an  example  of  parameter learning  by  VT  followed  by  the
computation  of  the Viterbi  explanation  for  a  program that  violates  the
exclusiveness  condition.  Fig.~\ref{prog:problog_graph}  is  a PRISM  program
translated from a ProbLog program\footnote{
The     program     is     taken     from     the     tutorial     at     {\tt
  http://dtai.cs.kuleuven.be/problog/}.
} that  computes a path  between two nodes  (and its probability) in  a graph.
The graph has six nodes. Edges  are assigned probabilities and we express this
fact  by  attaching  an  {\tt   msw}  atom  to  an  atom  {\tt  d\_e($x$,$y$)}
representing an edge $x - y$ in the program.  For example (directed) edge {\tt
  d\_e(1,2)} between  node 1 and node  2 is assigned probability  {\tt 0.9} as
indicated  by {\tt  msw(d\_e(1,2),on)}  following its  value declaration  {\tt
  values(d\_e(1,2),[on,off],[0.9,0.1])} in the program.

Observe  that  a  ground  top-goal  {\tt  path(X,Y)} causes  a  call  to  {\tt
  d\_e(X,Z)} with  {\tt X} ground  and {\tt Z}  free that calls more  than one
clause,  which  leads  to   the  violation  of  the  exclusiveness  condition.
Nonetheless we can learn parameters  by VT and compute the Viterbi explanation
for this  program.

Fig.~\ref{prog:sample_session}   is   a  sample   session   doing  this.    In
Fig.~\ref{prog:sample_session}, we  first compute  the Viterbi path  {\tt VE},
i.e.\ the  most probable path  between node 1  and node 4 and  its probability
{\tt  P} by applying  the built-in  predicate {\tt  viterbif/3} to  goal {\tt
  path(1,4)}\footnote{
{\tt viterbif(path(1,4),P,\_X)}  returns respectively the  Viterbi explanation
in {\tt \_X} and its  probability in {\tt P}.  {\tt viterbi\_switches(\_X,VE)}
extracts the Viterbi path {\tt VE} as a conjunction of {\tt msw} atoms.
}.  We  next renew parameters  by learning them  using VT from  observed goals
{\tt path(1,4),path(1,3)...}\footnote{
{\tt set\_prism\_flag(learn\_mode,ml\_vt)}  tells the  PRISM system to  use VT
when {\tt learn/1} is invoked.
} 
\begin{figure}[b]
\rule{1.0\hsize}{0.8pt}
\begin{verbatim}
?- viterbif(path(1,4),P,_X),viterbi_switches(_X,VE)
P = 0.432
VE = [msw(d_e(1,2),on),msw(d_e(2,3),on),msw(d_e(3,4),on)]

?- set_prism_flag(learn_mode,ml_vt).

?- learn([path(1,4),path(1,3),path(2,4),path(2,5),path(3,6)]).
...

?- viterbif(path(1,4),P,_X),viterbi_switches(_X,VE)
P = 0.104
VE = [msw(d_e(1,6),on),msw(d_e(6,5),on),msw(d_e(5,4),on)]
\end{verbatim}
\rule{1.0\hsize}{0.8pt}
\caption{ A sample session }
\label{prog:sample_session}
\end{figure}
Finally  we compute  the  Viterbi path  again  that is  determined by  learned
parameters and see  whether the learning changes the Viterbi  path or not.  In
the session, the Viterbi path changed after  learning from {\tt 1 -> 2 -> 3 ->
  4} to {\tt 1 -> 6 -> 5 -> 4} together with their probabilities from 0.432 to
0.161.

VT  thus  enables  us to  learn  parameters  from  programs that  violate  the
exclusiveness condition.  However we have to  recall at this point that VT has
an objective function different from likelihood and is biased \cite{Lember07},
and  the effect  of removing  the exclusiveness  condition on  the  quality of
parameters estimated by VT is unknown at the moment, which remains as a future
research topic.

\section{Related work and discussion} 
\label{sec:discussion}

VT  is  closely related  to  K-means  \cite{MacQueen67}  which is  a  standard
clustering method  for continuous data. If  we apply VT to  a Gaussian mixture
for clustering of  continuous data with an assumption of  a common variance to
all composite Gaussian distributions,  the resulting algorithm is identical to
K-means. In this sense, the usefulness  of VT is established.  Actually VT has
been             used              in             various             settings
\cite{Brown93,Juang90,Strom99,Joshi06,Spitkovsky10,Lomsadze05} and also in the
SRL frameworks  that deal  with structured data  \cite{Singla05,Huynh10} where
the  algorithmic essence  of VT,  coordinate ascent  on parameters  and target
variables with argmax  operation applied to the latter, is used.

Despite  its   popularity  however,  it  seems   that  VT  so   far  has  been
model-specific and  only model-specific  VT algorithms have  been implemented.
In this paper  we gave a unified  treatment to VT for discrete  models for the
first time to our knowledge, and derived the VT algorithm for PRISM which is a
single generic algorithm applicable to any discrete model as long as the model
is described by a  PRISM program.  Since our derivation of VT  is based on the
reduction of goals  to AND-OR propositional formulas, it  seems quite possible
for  other  logic-based modeling  languages  that  use  BDDs such  as  ProbLog
\cite{DeRaedt07,Kimmig08}  and  PITA \cite{Riguzzi11}  to  introduce  VT as  a
parameter learning routine.

One of the unique features of VT is its affinity with discriminative modeling.
Write the VT's objective function $L_{VT}(y \mid \theta)$ as follows.
\begin{eqnarray*}
L_{VT}(y  \mid  \theta) 
     & = &  \log\;  p(x^*,y \mid \theta)p(\theta) \\
x^*  & = & {\rm argmax}_{x}\; p(x,y \mid \theta)p(\theta) \\
    & = &  {\rm argmax}_{x}\; p(x \mid y,\theta).
\end{eqnarray*}
This  means that although  PRISM is  intended for  generative modeling,  VT in
PRISM  computes   the  Viterbi  explanation  $x^*$  that   gives  the  highest
conditional  probability  $p(x^*  \mid  y,  \theta)$ for  $y$  whose  form  is
identical to the objective function in discriminative modeling and the Viterbi
explanation is chosen in the same way as discriminative modeling does provided
the hidden  variable is a  prediction target.  When  this condition is  met VT
shows   good    performance   as   demonstrated   by    the   experiments   in
Section~\ref{sec:grammar} but if not, VT does not necessarily outperform other
parameter  learning  methods  as  exemplified  in  Section~\ref{sec:nbh}.   It
therefore seems  reasonable to say that  VT is effective  for prediction tasks
when the prediction target coincides  with hidden variables in a probabilistic
model, though we obviously need more experiments.

As  a coordinate ascent  local hill-climber,  VT is  sensitive to  the initial
parameters and  also sensitive  to the Viterbi  explanation.  To  mitigate the
sensitivity problem with initial parameters, we used 50 time random restart in
the  learning  experiments in  Section~\ref{sec:grammar}.   To  cope with  the
sensitivity  to  the  Viterbi  explanation,  it is  interesting  to  introduce
$k$-best explanations as discussed in \cite{Gutmann08} and replace the Viterbi
explanation in  VT with  them.  This  approach will give  us control  over the
sensitivity and computation time by  choosing $k$ and seems not very difficult
to implement  in PRISM  as $k$-best  explanations for a  goal $G$  are already
computed by built-in predicates such as {\tt n\_viterbi($k$,$G$)}.

Since  VT in  PRISM  runs on  explanation  graphs obtained  from all  solution
search, it requires  time for all solution search (by  tabling) and also space
to store discovered explanation graphs.  It is possible, however, to implement
VT without  explanation graphs, and to  realize much more memory  saving VT by
repeating search for a Viterbi explanation  in each cycle of VT.  We note this
approach particularly fits well  with mode-directed tabling \cite{Zhou10}.  In
mode-directed  tabling, we  can search  for partial  Viterbi  explanations for
subgoals  efficiently without  constructing  explanation graphs  and put  them
together to form a larger Viterbi explanation for the goal.  Currently however
mode-directed  tabling  is  not  available   in  PRISM.   We  are  planing  to
incorporate it in PRISM in the near future.

\section{Conclusion} 
\label{sec:conclusion}
We introduced VT (Viterbi training)  to PRISM to enhance PRISM's probabilistic
modeling power.  PRISM becomes the first SRL (statistical relational learning)
language  \cite{Getoor07,DeRaedt08b} in  which VT  is available  for parameter
learning to our knowledge.

Although  VT has  already  been used  in  various models  under various  names
\cite{Brown93,Juang90,Strom99,Joshi06,Spitkovsky10,Lomsadze05},  we  made  the
following contributions to VT.  One  is a generalization by deriving a generic
VT algorithm for PRISM, thereby making  it uniformly applicable to a very wide
class  of discrete  models described  by PRISM  programs ranging  from  BNs to
probabilistic grammars.  The other is an empirical clarification of conditions
under which VT  performs well.  We conducted learning  experiments with a PCFG
and a PLCG using VT  and confirmed VT's excellent parsing performance compared
to  EM, MAP and  VB.  We  also conducted  a learning  experiment with  NBH for
classification tasks.   Putting the results of these  experiments together, we
may say that VT performs well when hidden variables are a prediction target.

From  the viewpoint  of PRISM,  VT improves  PRISM first  by  realizing faster
convergence  compared to EM,  second by  providing the  user with  a parameter
learning method  that can learn parameters  good for prediction,  and third by
providing  a solution  to  the  problem of  the  exclusiveness condition  that
hinders PRISM  programming.  Thanks to  VT, we are  now able to  use arbitrary
programs with inclusive-or for probabilistic modeling.

Last but not least we can say that as VT in PRISM is general and applicable to
any  PRISM program, it  largely reduces  the need  for the  user to  develop a
specific VT algorithm for a specific model.  Furthermore since VT in PRISM can
be  used just  by  setting a  PRISM  flag appropriately,  it  makes VT  easily
accessible to (probabilistic) logic programmers.


%


\end{document}